%% file: main.tex
\documentclass[10pt,twocolumn,letterpaper]{article}

\usepackage{cvpr}      % To produce the REVIEW version

\input{preamble}
\definecolor{cvprblue}{rgb}{0.21,0.49,0.74}
\usepackage[pagebackref,breaklinks,colorlinks,allcolors=cvprblue]{hyperref}
\usepackage{microtype} % 导言区添加
\usepackage{ragged2e}
\microtypesetup{protrusion=true, expansion=true}
\usepackage{graphicx}
\usepackage{caption}
\usepackage{makecell}
\usepackage[misc]{ifsym}
\def\paperID{} % *** Enter the Paper ID here
\def\confName{CVPR}
\def\confYear{2026}

\title{Premier: Personalized Preference Modulation with Learnable User Embedding in Text-to-Image Generation}
\author{Zihao Wang$^1$ \ Yuxiang Wei$^1$ \ Xinpeng Zhou$^1$ \ Tianyu Zhang$^2$ \  Tao Liang$^2$ \\
Yalong Bai$^2$ \ Hongzhi Zhang$^{1}$\textsuperscript{(\Letter)} \  Wangmeng Zuo$^1$\\ \\
$^1$Harbin Institute of Technology \quad $^2$ Duxiaoman  \\
}

\begin{document}

\twocolumn[{
    \maketitle
    \vspace{-2.5em}
    \begin{center}
        \includegraphics[width=\linewidth]{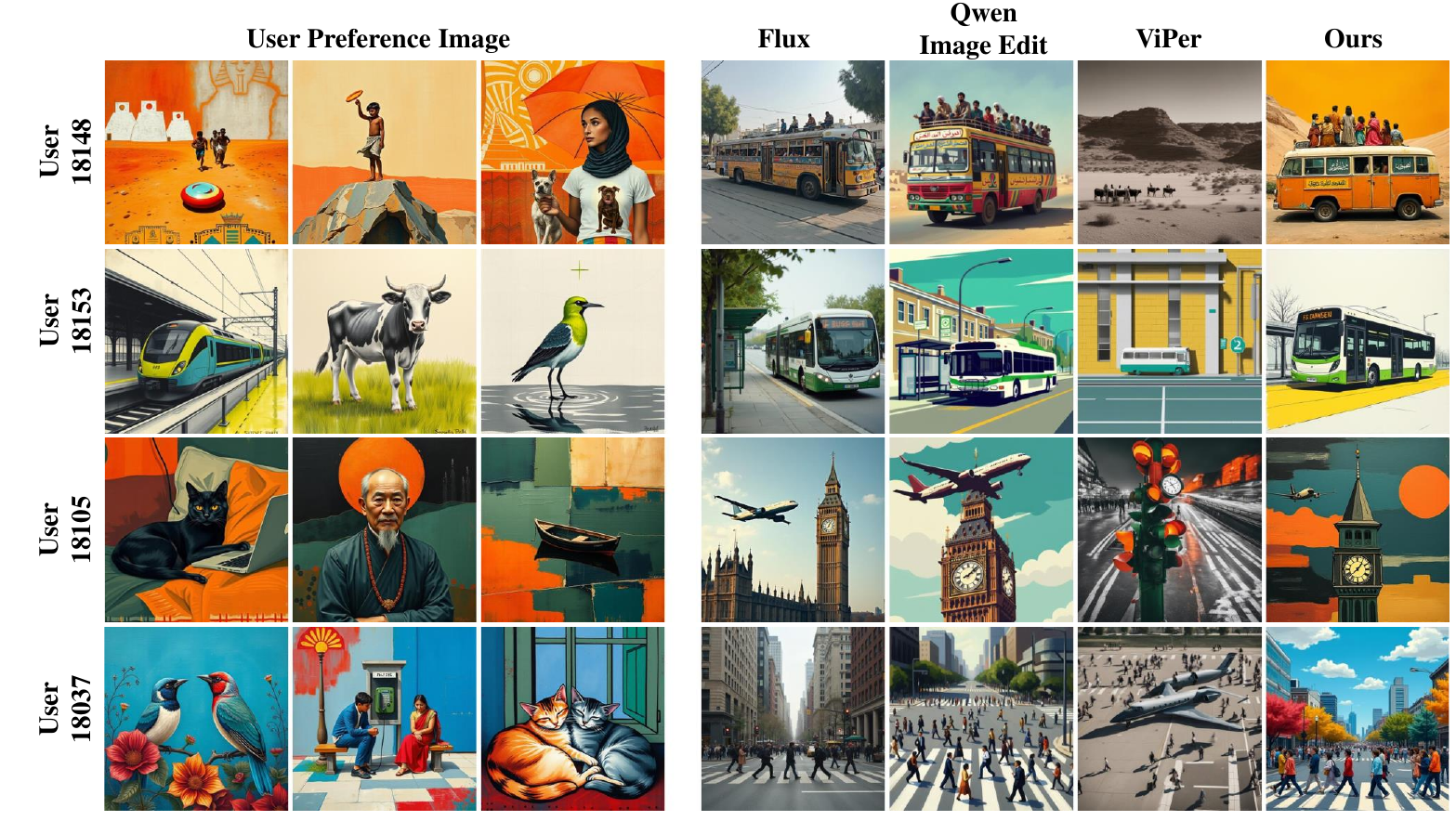}
        \captionof{figure}{In our approach, user preference descriptions are not required and only user-provided preference images are needed. A learnable user embedding is obtained by training on these images, and this embedding accurately captures the user’s preference information. }
        \label{fig:teaser}
    \end{center}
}]

\input{sec/0_abstract}    
\input{sec/1_intro}
\input{sec/2_related}
\input{sec/3_preliminaries}
\input{sec/4_method}
\input{sec/5_experiments}
\input{sec/6_conclusion}

\input{sec/X_suppl}
\small
\bibliographystyle{ieeenat_fullname}
\bibliography{main}

\end{document}

%% file: sec/0_abstract.tex
\begin{abstract}
    % Text-to-image generation technology has advanced rapidly, yet it still struggles to accommodate the diverse preferences of individual users. Previous methods extract user preferences using multimodal models, but the resulting natural language descriptions or hidden states often suffer from poor adherence by the generative model. To address this, we propose Premier that leverages learnable user preference embeddings and preference adapters for personalized preference image generation. 
    % To produce more accurate and context-aware modulation directions, we design prompt-preference modulation, which enables interaction between user embedding and input context.
    % Furthermore, our method employs a dispersion loss to encourage clearer separation among different users' embeddings in the feature space, thus promoting better alignment between the generated images and various preferences. 
    % To address the challenge of limited user-provided preference images, we propose training new user preferences by representing them as linear combinations of user preference embeddings from the training set. 
    % Experiments show that our method achieves superior preference alignment compared to other approaches with the same history length, demonstrating strong performance in text consistency, ViPer proxy scoring, and human expert evaluation.

Text-to-image generation has advanced rapidly, yet it still struggles to capture the nuanced user preferences. 
Existing approaches typically rely on multimodal large language models to infer user preferences, but the derived prompts or latent codes rarely reflect them faithfully, leading to suboptimal personalization.
We present Premier, a novel preference modulation framework for personalized image generation. 
Premier represents each user's preference as a learnable embedding and introduces a preference adapter that fuses the user embedding with the text prompt.
To enable accurate and fine-grained preference control, the fused preference embedding is further used to modulate the generative process.
To enhance the distinctness of individual preference and improve alignment between outputs and user-specific styles, we incorporate a dispersion loss that enforces separation among user embeddings. 
When user data are scarce, new users are represented as linear combinations of existing preference embeddings learned during training, enabling effective generalization.
Experiments show that Premier outperforms prior methods under the same history length, achieving stronger preference alignment and superior performance on text consistency, ViPer proxy metrics, and expert evaluations. The code is available at \url{https://github.com/120L020904/Premier}.
    
\end{abstract}

%% file: sec/1_intro.tex
\section{Introduction}
\label{sec:intro}

Diffusion models ~\cite{ho2020denoising,liu2023flow,song2020score} have significantly improved the quality of image generation.
An increasing number of users are using text-to-image models~\cite{wu2025qwenimagetechnicalreport,labs2025flux1kontextflowmatching} to generate images.
There are many non-professional users, who often find it difficult to accurately describe the images they want~\cite{Prompist23nips, ma2024exploring,gao2025devil,mo2024dynamic}.
Moreover, image preferences are sometimes difficult to describe clearly in words.
This makes it challenging for current image generation models to cater to the diverse preferences of various users.
%
%Although users may encounter challenges in describing their preferences in words, they can select images they like.
Although textual descriptions can be challenging to articulate, users can express their preferences through actions such as clicking, downloading, or favoring images. 
%
%These selected images contain valuable information about the user's preferences.
In practical applications or within user communities, the images selected by users inherently contain valuable information about their preferences.
%
% This paper proposes leveraging learnable user embedding to extract preference information from users' preference images.
This paper aims to extract preference information from their selected images and then leverage these features in the image generation process to produce images that better align with their individual preferences.

The key challenge in user preference generation lies in how to represent user preferences and how to use them to control image generation. 
Current methods ~\cite{shen2024pmg,bian2025icg,ling2025ragar,xu2025drc,xu2025personalized, li2025instant} tend to rely on large multimodal models to extract user preferences from preference images, which introduces several limitations.
One strategy involves extracting the hidden states from a multimodal model as the user preference representation and injecting this preference into the image generation process via a connector module.
However, the connector module can become a performance bottleneck, potentially degrading the rich user preference information extracted by the large multimodal model.
Another approach directly leverages large multimodal models to extract semantic preferences in the form of natural language descriptions.
However, text-to-image models may struggle with instruction-following when faced with complex or nuanced preference descriptions.
Additionally, the weak correlation in user preference history further complicates preference extraction.
As the history length increases, large multimodal models tend to overlook fine-grained distinctions across individual samples, undermining the fidelity of the extracted preferences.

These challenges motivate us to explore whether learnable embeddings trained via diffusion loss backpropagation can serve as more effective preference representations.
In the context of condition injection within the multimodal diffusion transformer (MM-DiT) ~\cite{peebles2023scalable}, current methods primarily concatenate conditional tokens with generated image tokens~\cite{tan2025ominicontrol, xiao2025omnigen, wang2024easycontrol}.
Zhang et al ~\cite{lv2025rethinking} point out that concatenating conditional tokens for condition control suffers from the token dilution problem.
Specifically, since text and image sequences typically consist of a large number of tokens, effective control of image generation by the user embedding would require either increasing the number of dedicated tokens or fine-tuning via LoRA, both of which may compromise the model’s original performance.
In contrast, the modulation approach enables preference modulation at the level of each individual text token, thereby avoiding the aforementioned token dilution issue.
Moreover, since the modulation process directly takes text tokens from the text encoder as input, it can be applied prior to the image generation process.
Therefore, our method focuses on adding preference conditions via modulation, which is a more flexible and adaptive approach that better suits the demands of personalized preference generation.

\begin{figure*}[!t]
    \centering
    \includegraphics[width=\linewidth]{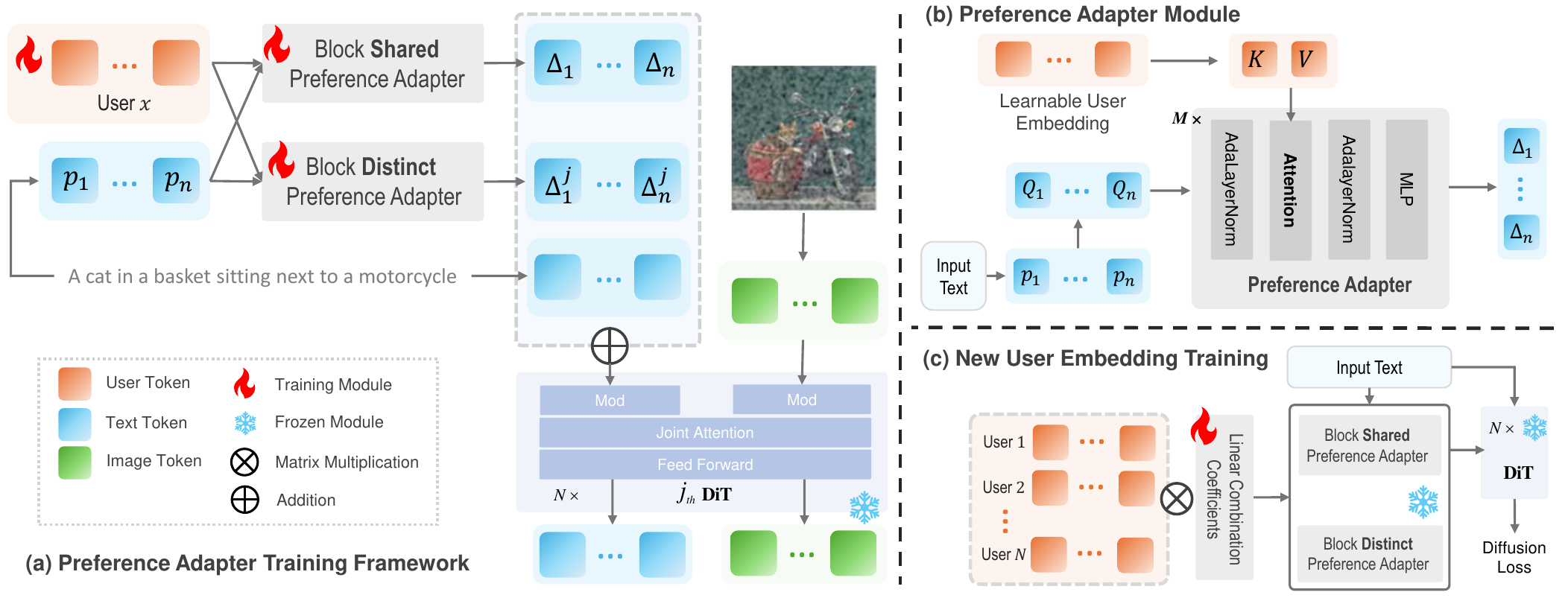}
    \caption{\textbf{Premier training framework.} (a) During the training of the preference adapters, the user preference embeddings and the adapters are jointly optimized. The block-shared adapter produces a uniform modulation direction across all DiT blocks, whereas the block-distinct adapter generates different modulation directions for different DiT blocks.  (b) Each preference adapter takes the learnable user embedding and the input text tokens as inputs, and outputs a preference modulation direction for every text token, enabling fine-grained and context-aware modulation. (c) Our method obtains the new user’s preference embedding as a linear combination of training-set user preference embeddings. During this stage, only the linear combination coefficients are optimized. This strategy yields a more stable user preference embedding when the user’s historical data is limited.}
    \label{fig:method}
\end{figure*}
We propose Premier (\underline{P}ersonalized P\underline{re}ference \underline{M}odulat\underline{i}on with L\underline{e}arnable Use\underline{r} Embedding in Text-to-Image Generation), which leverages learnable user embeddings to represent user preferences. 
During training, user embeddings can aggregate information across users' preference images.
Additionally, we introduce a preference adapter to enable interaction between the user embedding and the input prompt, allowing the preference to be expressed adaptively under different text tokens.
The preference adapter takes text tokens and the user embedding as inputs, and outputs a preference modulation direction for each text token.
The modulation direction is added to the modulation vector in MM-DiT.
During practical training, we observed that the output of the preference adapter is not sufficiently sensitive to variations in user features.
Directly using the diffusion loss for alignment causes the preference adapter to overfit to text tokens, resulting in similar generated images across different users.
%reduced variability across users. %, i.e., generated preference images exhibit minimal distinctions between users.
%
To mitigate this issue, we introduce a dispersion loss that promotes separation between the adapter's output representations in the feature space.

%

%When users first start using the system, they can only provide a limited number of preference images.
Our framework also incorporates a cold-start strategy to handle new users with an extremely limited number of samples.
Directly training on sparse data leads to overfitting and unstable preference alignment.
We propose representing new users as linear combinations of user embeddings from the training set to obtain more robust embeddings.
%
%Since users in the training set have more abundant data, their embeddings are better learned, and their linear combinations can more effectively approximate the preferences of new users with sparse preference data.
%
In summary, our key contributions are as follows:

\begin{itemize}
    \item We employ learnable user embeddings to capture user preferences during training and utilize prompt preference modulation to enable interaction between the user embedding and input text, thereby producing more precise and context-aware modulation directions.

    \item  We introduce a dispersion loss to encourage clearer separation among modulation directions under different user embeddings, thereby promoting the generation model to produce images that are more precisely aligned with individual user preferences.

%We introduce a dispersion loss to encourage clearer separation among modulation directions under different user embeddings, thereby promoting more accurate preference alignment.

    \item We obtain the preference embedding for new users with limited preference data via a linear combination of training-set user preference embeddings, enhancing the stability of preference alignment. % With limited user preference data, training the linear combination coefficients yields better performance than directly optimizing the embedding.
\end{itemize}

%% file: sec/2_related.tex
\section{Related Work}
\label{sec:related}

%-------------------------------------------------------------------------
\subsection{Personalized  Image  Generation}
Personalized image  generation~\cite{li2025instant,kim2025draw,dang2025personalized,wallace2024diffusion,bian2025icg,xu2025drc,salehi2024viper,ling2025ragar,xu2025personalized,nabati2024preference,guo2025imagegem,dunloppersonalized,anonymous2025prefgen} extracts a user's image preferences from their preference data.
One approach is to extract preference information from users' historical text.
Tailored Visions~\cite{chen2024tailored} leverages a large language model to extract preference information from the user's historical input text and rewrites the user's current input text accordingly.
DrUM ~\cite{kim2025draw} leverages the user's historical input text and fuses it with the current input text in the feature space to produce tokens that integrate the user's preferences.
%
% User preferences sometimes are inherently difficult to articulate in natural language.
% %
% Due to a lack of expertise, users’ textual descriptions may diverge from their actual preference images—further limiting the applicability of the aforementioned methods.
%
Another line of work focuses on extracting user preferences from images and users' historical text.
PMG~\cite{shen2024pmg} combines soft preferences from the LLM's continuous feature space with hard preference text directly extracted by the LLM, jointly serving as the user's preference representation.
ViPer~\cite{salehi2024viper} extracts multi-dimensional user preferences from preferred images, non-preferred images, and user comments, and uses these multi-dimensional preferences to guide personalized image generation.
%
% Due to information loss in transferring knowledge from large multimodal models to image generation models, and the difficulty of image generators in accurately following complex instructions, these methods struggle to extract precise user preferences.
%-------------------------------------------------------------------------
\begin{figure*}[!t]
    \centering
    \includegraphics[width=\linewidth]{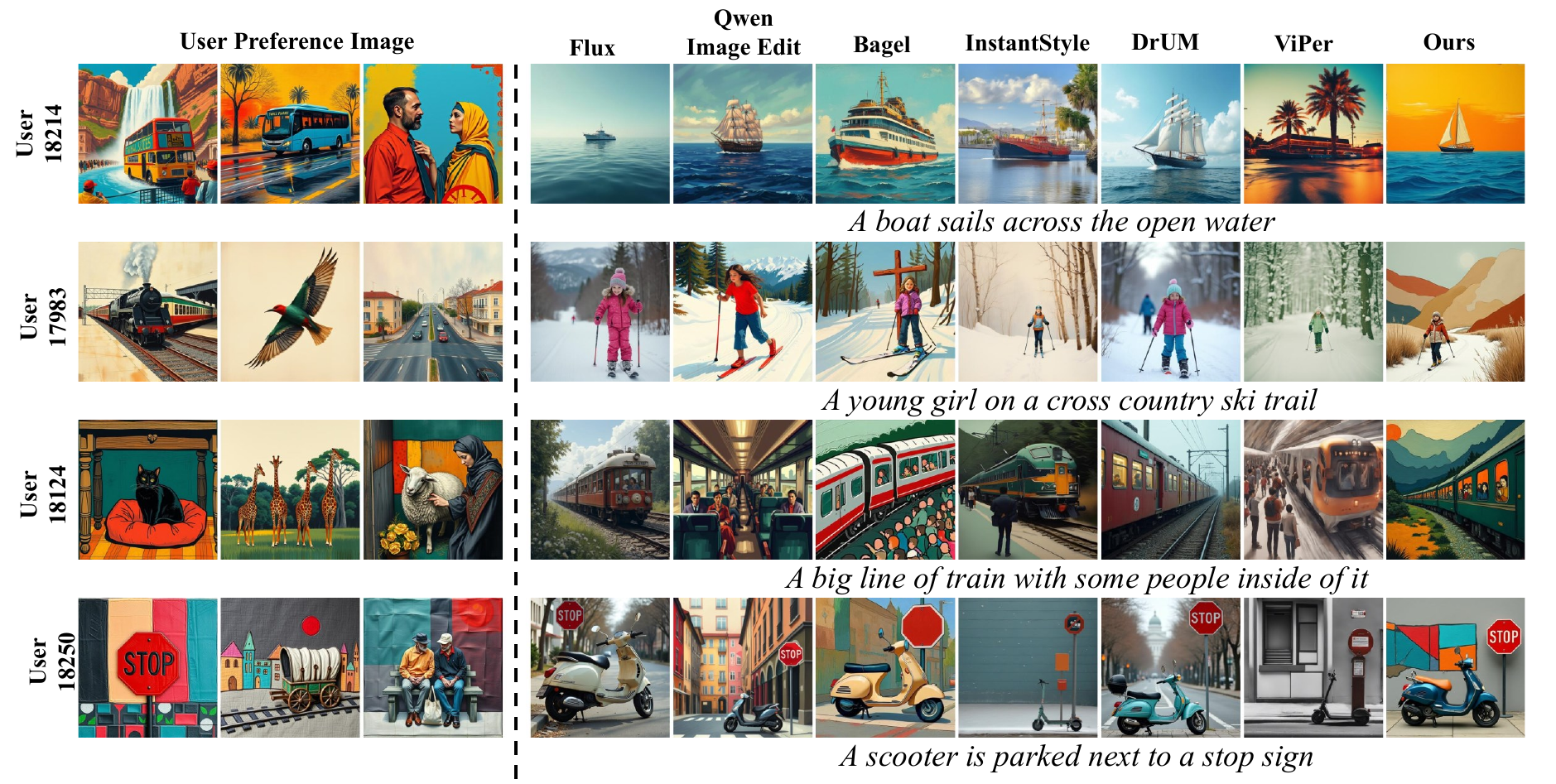}
    \caption{\textbf{Qualitative comparisons of Preference Alignment.} We compare the performance of our method with other approaches in user preference-aware image generation. The images generated by our method are closest to the user’s preferences while remaining faithful to the user-provided text prompt. }
    \label{fig:result_compare}
\end{figure*}

\subsection{Reference-based Image Generation}

Reference-based image generation ~\cite{ye2023ip,zong2024easyref,wang2024instantstyle,zhang2023adding,brack2024ledits++,wang2025ace,garibi2025tokenverse,chen2025xverse,tan2025ominicontrol,xiao2025omnigen,wang2024easycontrol,wang2025omnistyle,li2024styletokenizer,wang2025styleadapter} takes reference images as input conditions and jointly uses them with the user’s text prompt to control the generation of new images.
In models based on the UNet architecture, reference image guided generation is commonly realized via mechanisms such as IP-Adapter~\cite{ye2023ip} which introduces a dedicated trainable cross attention module to inject reference derived conditional features into the denoising backbone.
InstantStyle~\cite{wang2024instantstyle} injects style information from the conditioning image into the generation process through a style adapter.
EasyRef~\cite{zong2024easyref} leverages a Large Multimodal Model (LMM) to extract information from the conditioning image and uses it to control image generation.
In the MM-DiT architecture, there are two primary approaches to implement conditional control.
One concatenates conditional tokens with the image tokens to be generated, and the other introduces modulation directions prior to adaptive layer normalization ~\cite{xu2019understanding}(AdaLN).
With respect to conditional token concatenation, OmniGen~\cite{xiao2025omnigen} employs a hybrid attention mechanism that combines causal attention for text tokens and bidirectional attention for image tokens during cross-modal attention computation.
OminiControl~\cite{tan2025ominicontrol} achieves conditional control by training LoRA~\cite{hu2022lora} modules on the conditional control flow.
Regarding the incorporation of modulation directions, TokenVerse~\cite{garibi2025tokenverse} observes that injecting an offset direction into the text token modulation process within MM-DiT can effectively incorporate reference image information into the generated image.
XVerse~\cite{chen2025xverse} employs a resampler to enable interaction between image and text information, producing a modulation direction as output.
%
% Unified generation-and-understanding models such as Bagel~\cite{deng2025emerging} leveraging the comprehension capability of multimodal models to assist reference-guided image generation.
%-------------------------------------------------------------------------

%% file: sec/3_preliminaries.tex
\section{Preliminaries}

\subsection{Diffusion Models}
Most current image generation models are based on diffusion model, and modern diffusion models employ the flow matching~\cite{liu2023flow}. 
The flow matching approach generates supervision data for the diffusion steps via interpolation, formulated as:
\begin{equation}
    z_t=(1-t)\cdot z_0 + t\cdot z_1,
\end{equation}
where $z_0$ is the data sample. $z_1$ denotes the sample from the initial distribution, typically $z_1 \sim \mathcal{N}(0,1)$. $t \in [0,1]$ denotes the continuous time step. The model is trained to predict the conditional velocity $\mathbf{v}_\theta(\mathbf{z}_t, c,t)$:
\begin{equation}
    \mathcal{L}_\text{flow} = \mathbb{E}_{z_0,z_1,t} \left[ \left \| \mathbf{v}_\theta(\mathbf{z}_t, c,t) - (\mathbf{z}_1 - \mathbf{z}_0)\right\|_2^2 \right].
\end{equation}
This loss encourages the model to learn the transport path from noise to data, enabling fast and stable sampling.
During the reverse process, the model generates realistic images by solving the corresponding ordinary differential equation (ODE).

\subsection{Modulation in MM-DiT}
In MM-DiT, conditional information such as timestep is injected into both text and image tokens using a modulation mechanism.
The modulation mechanism uses MLP to synthesize a modulation vector $y$ from conditional inputs such as timesteps and text embeddings. Specifically:
\begin{equation}
    y = \mathcal{M}_p(\text{CLIP}(p))+\mathcal{M}_t(t),
\end{equation}
where $\mathcal{M}_p$ and $\mathcal{M}_t$ are distinct MLPs, applied separately to the CLIP~\cite{radford2021learning} text embedding and the timestep embedding respectively. $p$ denotes the text input provided by the user.
The modulation vector $y$ is passed through linear layers to generate modulation parameters for both text tokens and image tokens.
The modulation mechanism employs AdaLN and residual connections to fuse modulation parameters with image and text token.
Previous studies~\cite{garibi2025tokenverse,chen2025xverse,zhong2025mod} have found that adding a specific modulation direction $\Delta$ to the modulation vector can steer the object attributes corresponding to the modulated text tokens:
\begin{equation}
    y'_i = y +\Delta_i,
\end{equation}
where \( i \) denotes the index of the text token. 
In the original model, the modulation vector is shared across all tokens. 
In practice, this shared modulation vector is duplicated, and distinct modulation directions are added to produce token-specific modulation vectors for different tokens.
These modulation vectors endow the corresponding generated objects with attributes from the reference image.
%
% Compared to token concatenation, this conditional injection approach prevents the conditioning signal from being overwhelmed by the considerably larger number of image and text tokens, thereby enabling more effective preference conditioning.
\begin{figure}[!t]
    \centering
    \includegraphics[width=\linewidth]{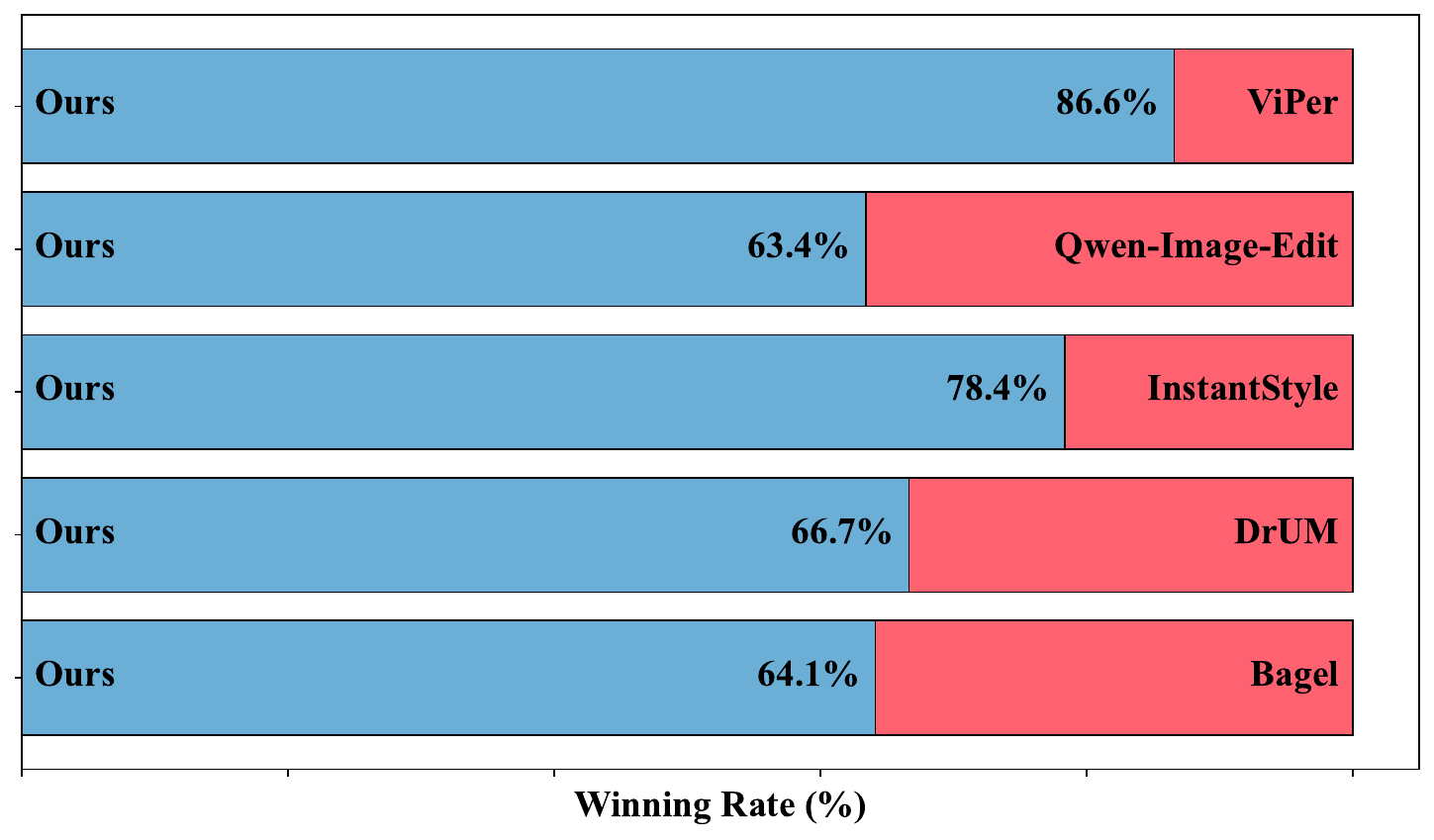}
    \caption{\textbf{User study results of our method compared with other methods.} Each human expert is presented with six historical preference images from the user, along with image pairs generated by our method and other baselines under the same text prompt. Experts are asked to select the image that best aligns with both the user’s preferences and the input text.}
    \label{fig:human_evaluation}
\end{figure}

%% file: sec/4_method.tex
\section{Proposed Method}

Our method uses learnable user embeddings to represent user preferences.
By training on users’ historical preference images, the learnable user embedding can be better adapted to text-to-image models and more accurately encode the users' preferences.
Our method uses preference adapters to fuse the learnable user embedding and the user's input text prompt, outputting a preference-aware modulation direction for each text token.
To enhance the discriminability among user preference embeddings and achieve better preference alignment, we introduce a dispersion loss.
This loss encourages the preference modulation directions of different users to be well-separated in the feature space.
To improve the robustness of user preference embeddings when user data is limited, we propose constructing user embeddings via linear combinations of embeddings from users in the training set.
This strategy better leverages the stable user preference embeddings from the training set, yielding a more robust user embedding especially when the user’s own preference data is scarce.

\subsection{Prompt Preference Modulation}

Current preference generation approaches often rely on multimodal models to extract user preferences.
However, the hidden states produced by large multimodal models differ in representation space and semantics from those used in text-to-image models, leading to information loss during the feature adaptation or transformation process.
Moreover, when relying on natural language descriptions provided by multimodal models, text-to-image models often struggle with instruction following, leading to inaccurate generation of preference-aligned images.

Our method more accurately represents user preferences through learnable user embeddings during training.
To enable flexible preference conditioning, our method integrates user preferences through modulation.
Additionally, to achieve fine-grained preference modulation at the token level, our method employs two preference adapters that model the interaction between the user preference embedding $e_u$ and each text token embedding $e_{p_i}$ via cross-attention.
These two preference adapters produce two types of modulation directions to each text token:
\begin{equation}
    y_i^{j} = y + \Delta_\text{shared}(e_u, e_{p_i}) + \Delta^j_\text{disdinct}(e_u, e_{p_i}),
\end{equation}
where $\Delta_\text{shared}(e_u, e_{p_i})$ is identical across all DiT blocks.
$\Delta^j_\text{distinct}(e_u, e_{p_i})$ differs across DiT blocks and $j$ is the index of DiT block.
As illustrated in ~\cref{fig:method}, the two preference adapters employ a cross-attention mechanism, using the user-provided text as queries (Q) and the user preference embedding as keys (K) and values (V), to output a token-specific modulation direction for each text token.
Compared to the block-shared preference adapter, the block-distinct preference adapter expands the output dimension in the final layer, enabling it to produce different modulation directions for distinct DiT blocks.

\begin{figure*}[!t]
    \centering
    \includegraphics[width=\linewidth]{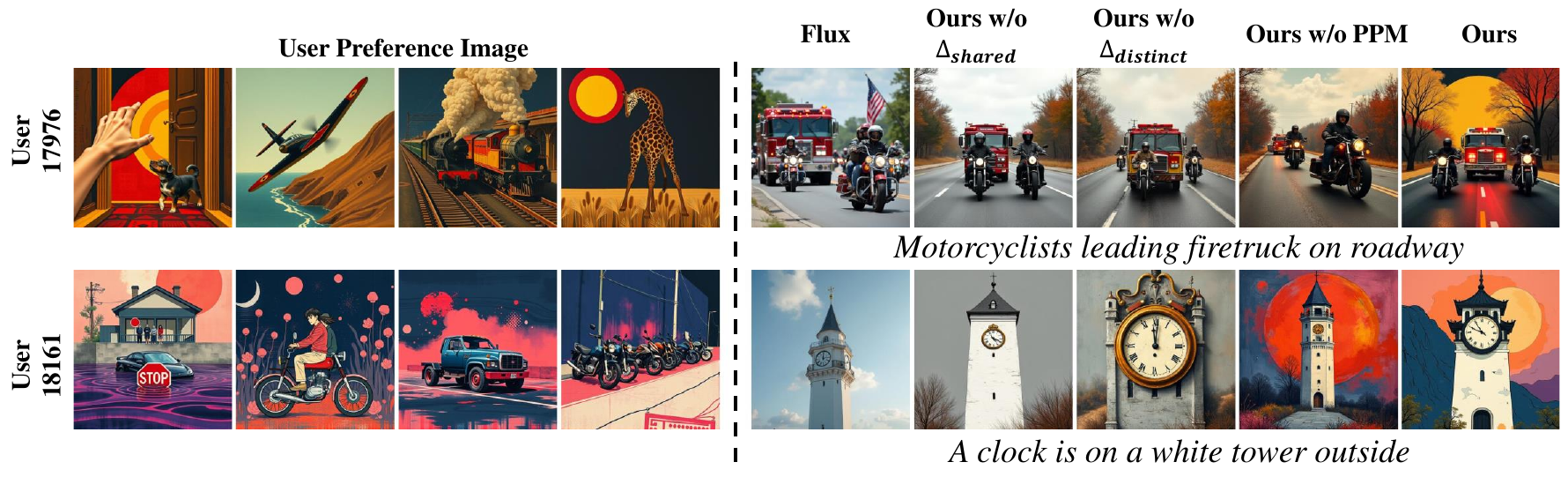}
    \caption{\textbf{Qualitative ablation comparison of our method.} Ablating either of the two preference adapters leads to a significant performance drop, confirming their necessity. Ablating the text-preference modulation also degrades user-preference-aware image generation.}
    \label{fig:result_abla}
\end{figure*}
\input{table/main_comparison}
\input{table/ablation}
\subsection{Learning Objective}
As shown in ~\cref{fig:method}, our method divides the training into two stages.
In the first stage, our method trains the preference adapters together with the user preference embeddings on the training set.
In the second stage, our method only trains the preference embedding for new users, while keeping the preference adapters frozen.

When relying solely on the $\mathcal{L}_\text{flow}$ for preference adapters training, we observe that the model learns only generic preferences and fails to effectively discriminate between different user preference embeddings.
This is caused by overfitting of the preference adapter to the text tokens.
We expect the preference modulation directions to be primarily distinguished based on user preference embeddings, and only subsequently refined at the level of individual text tokens.
Therefore, our method adopts the contrastive learning paradigm by treating modulation directions conditioned on other users’ preference embeddings within the same batch as negative samples.
Following the dispersion loss framework~\cite{wang2025diffuse}, the standard flow matching loss can be viewed as a positive-sample alignment term.
To encourage diversity and user-wise discrimination, we further introduce a dispersion loss based on InfoNCE~\cite{he2020momentum}:
\begin{equation}
    \mathcal{L}_\text{disp} = \log \sum_j \exp (-\mathcal{D} (\Delta_\theta(e_{u}, e_p),\Delta_\theta(e_{u'}, e_p))),
\end{equation}
where $\mathcal{D}$ denotes a distance function, and in our method, it is L2 norm.
And the input text $p$ is set to an empty text prompt.
$u$ and $u'$ denote two different user embeddings.
The two preference adapters compute their dispersion losses separately, each directly using its own outputs within a batch.
The final loss function is formulated as:
\begin{equation}
    \mathcal{L} =  \mathcal{L}_\text{flow}+\lambda_\text{shared}\mathcal{L^\text{shared}_\text{disp}}+\lambda_\text{distinct}\mathcal{L^\text{distinct}_\text{disp}}.
\end{equation}
During the training of the preference adapters, the loss function encourages them to produce distinct preference modulation directions under different user preference embeddings.
During new user training, the preference adapters have already been trained.
Therefore, only the flow matching loss is required.

\subsection{Preference Learning of New Users}

In practice, users initially provide only a limited number of preference images, which makes it challenging to train an accurate user preference embedding.
However, users in the training set all have abundant preference data, resulting in more stable and more reliable user preference embeddings.
Therefore, our method represents new users as linear combinations of training-set users to address the instability and overfitting of user preference embeddings trained on insufficient historical data for new users.
During training, we only optimize the coefficients of the linear combination, while keeping both the preference adapters and the training-set user preference embeddings frozen.
In this way, even with limited historical data, the new user embedding can leverage the well-trained embeddings from the training set to obtain a stable and meaningful representation.
Although directly training user preference embeddings performs well when abundant historical data is available, in real-world applications users often provide only a limited number of preference images.
Therefore, during training new user embedding, we adopt the strategy of optimizing linear combination coefficients.

%% file: table/main_comparison.tex
\begin{table*}[!ht]
    \centering
    \resizebox{\linewidth}{!}{
        \begin{tabular}{l|cccccccc}\hline
        ~ & Data & Flux~\cite{labs2025flux1kontextflowmatching} & InstantStyle~\cite{wang2024instantstyle} & Bagel~\cite{deng2025emerging} & Qwen Image Edit~\cite{wu2025qwenimagetechnicalreport} & DrUM~\cite{kim2025draw}  & ViPer~\cite{salehi2024viper}  & Ours \\ \hline
        ViPer Score$\uparrow$ & 0.8890 & 0.3953 & \underline{0.6277} & 0.5075 & 0.4688 & 0.4791 & 0.5159 & \textbf{0.6889} \\ 
        ViPer Rate$\uparrow$ & - & - & \underline{0.777} & 0.703 & 0.613 & 0.573 & 0.676 & \textbf{0.876} \\ 
        CLIP T2I$\uparrow$ & 0.3027 & 0.3089 & 0.2988 & 0.3107 & 0.3101 & \underline{0.3072} & 0.2981 & \textbf{0.3183} \\ 
        LPIPS$\downarrow$ & - & - & 0.6641 & 0.6438 & \underline{0.6407} & 0.6541 & 0.6564 & \textbf{0.5986} \\ \hline
    \end{tabular}
    }
    \caption{\textbf{Quantitative comparison of preference alignment in image generation across different methods.} The best results are highlighted in bold, while the second-best is underlined. DrUM, ViPer, and our method take 8 user historical data for preference generation, whereas Bagel, Qwen-Image-Edit, and InstantStyle use a single user history data. Under evaluation by the ViPer proxy model, our method achieves the highest alignment with user preferences, while also generating images with high text-image fidelity. }
    \label{tab:compare}
\end{table*}

%% file: table/ablation.tex
\begin{table}[!t]
    \centering
    \resizebox{\linewidth}{!}{
        \begin{tabular}{l|cccc}
        \hline
        ~ & ViPer Score $\uparrow$& ViPer Rate $\uparrow$& Clip Score T2I $\uparrow$& LPIPS $\downarrow$\\ \hline
        Ours w/o $\Delta_\text{shared}$ & 0.4818 & 0.667 & \underline{0.3162} & 0.6247 \\ 
        Ours w/o $\Delta_\text{distinct}$ & 0.4917 & 0.669 & 0.3131 & 0.6353 \\ 
        Ours w/o Disp Loss & 0.4498 & 0.618 & \underline{0.3162} & 0.6249 \\ 
        Ours w/o PPM & \underline{0.6492} & \underline{0.840} & 0.3074 & \underline{0.6225} \\ 
        Ours & \textbf{0.6889} & \textbf{0.876} & \textbf{0.3183} & \textbf{0.5986} \\  \hline
    \end{tabular}
    }
    
    \caption{\textbf{Quantitative evaluation of preference image generation after ablation.} The best results are highlighted in bold, while the second-best is underlined. Ablating the dispersion loss leads to a significant performance drop, while removing the modulation of prompt–preference interaction yields suboptimal results.}
    \label{tab:abla}
\end{table}

%% file: sec/5_experiments.tex
\section{Experiments}

In this section, we will first introduce the details of our implementation and evaluation.
Then, we compare our method against with existing approaches to demonstrate its effectiveness.
Finally, ablation studies have been conducted to analyze the effects of different user embedding training strategies and the amounts of user history.

\subsection{Experimental Details}

\noindent \textbf{Dataset.}
We train our model on the dataset PrefBench introduced by ~\cite{anonymous2025prefgen}, which includes multidimensional preference annotations and real human data.  
Specifically, we select 1,000 users as the training users. 
For each user, 40 $\sim$ 80 preference examples are available for training.
To evaluate our method, we additionally select 50 users outside the training set as test users. 
For each test user, we use 8 preference samples to learn the preference embedding and 20 samples as test data for evaluation.

\noindent \textbf{Implementation Details.}
The model is implemented based on open-sourced FLUX.1-dev and trained in two stages.
In the first stage, we jointly train the preference adapters and the preference embeddings for the training users.
We use the Prodigy optimizer~\cite{mishchenko2024prodigy} with a learning rate of 1.0, training for 4,000 epochs with a batch size of 16.
%
% In each epoch, the user preference embeddings of the training set are updated once.
%
$\lambda_\text{shared}$ and $\lambda_\text{distinct}$ are set to 0.1.
Each user's preference embedding is initialized as a tensor of shape \(30 \times 1024\).
Both preference adapters adopt a stack of three attention blocks, as illustrated in \cref{fig:method}.
The first training stage was conducted on 8 NVIDIA A800 GPUs.
In the second stage, we train preference embeddings for new users, which are represented as a linear combination of the user embeddings learned in the first stage. 
We use the Prodigy optimizer with a learning rate of 1.0, a batch size of 2, and 5,000 training steps. 
This stage is conducted on a single NVIDIA A800 GPU.

\begin{figure}[!t]
    \centering
    \includegraphics[width=\linewidth]{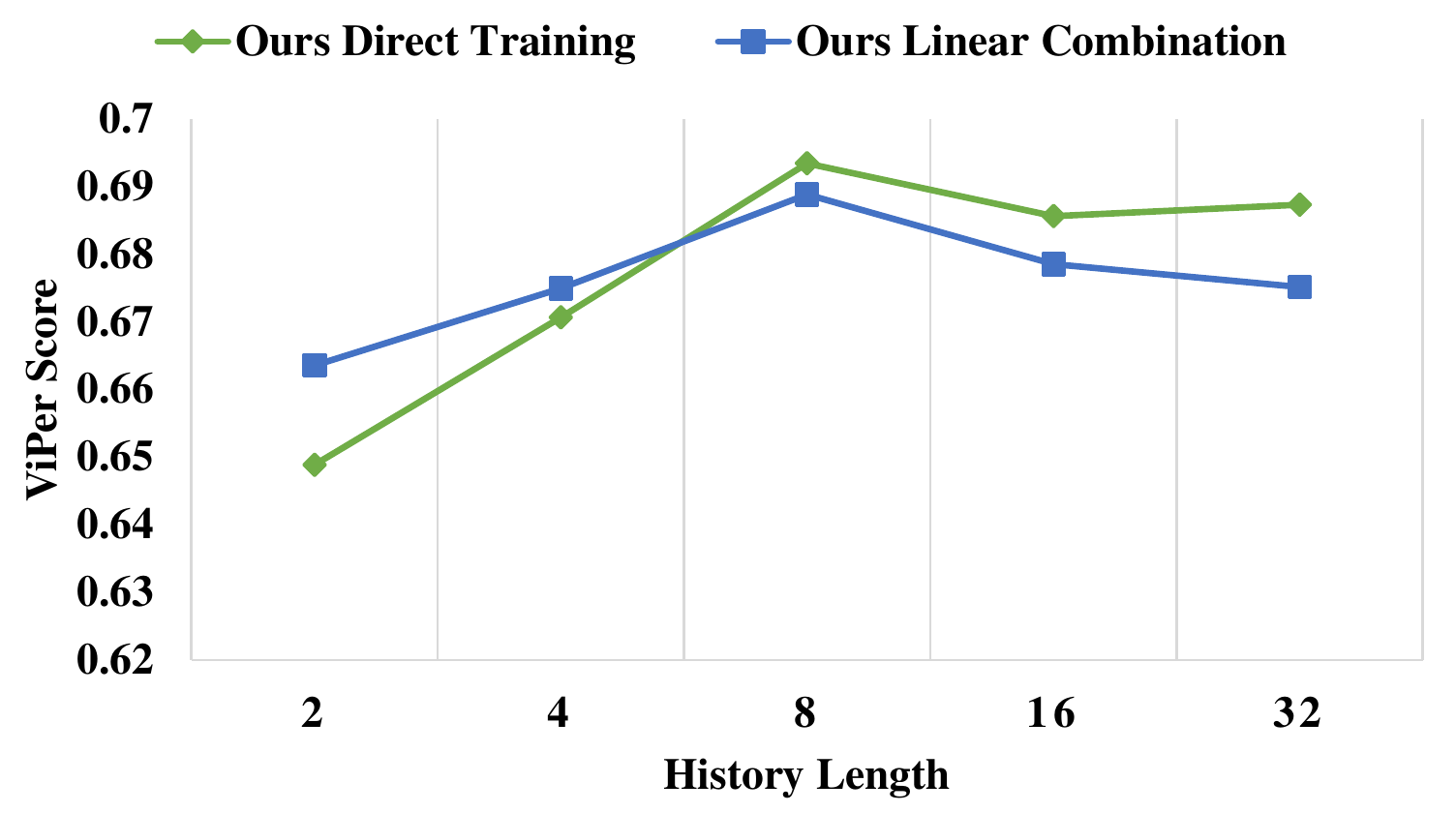}
    \caption{\textbf{The relationship between ViPer Score and user history length.} Our method’s strategy of training linear combination coefficients outperforms direct embedding training when the user history is limited.}
    \label{fig:history_viper_score}
\end{figure}
\begin{figure}[!t]
    \centering
    \includegraphics[width=\linewidth]{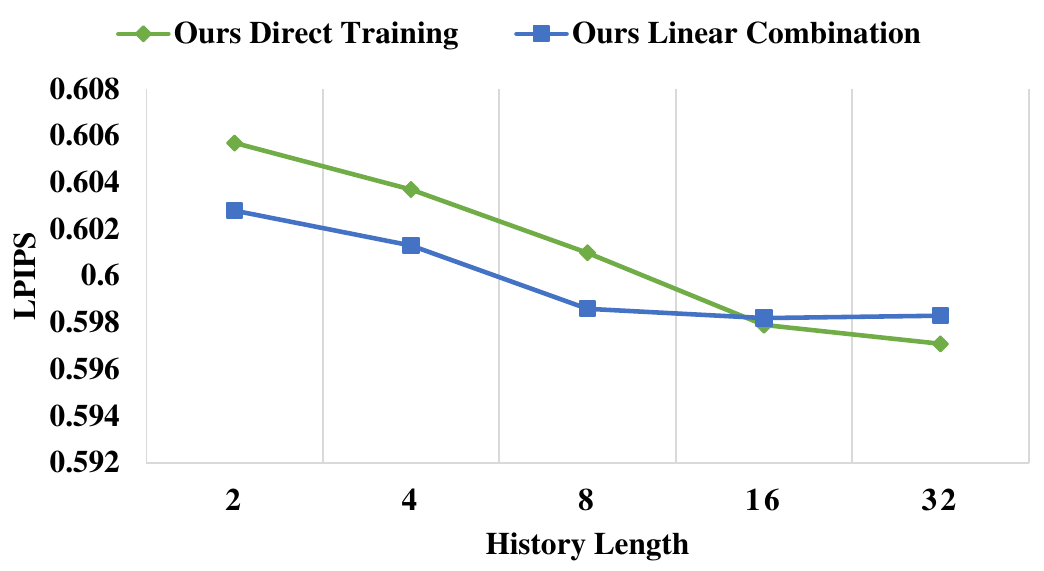}
    \caption{\textbf{The relationship between LPIPS and user history length.} Our linear combination approach demonstrates a more pronounced advantage in LPIPS, consistently outperforming direct training when the history size is up to 16 samples.}
    \label{fig:history_lpips}
\end{figure}

\noindent \textbf{Evaluation Metrics.}
To evaluate our method, we employ the ViPer score and win rate~\cite{salehi2024viper}, CLIP score~\cite{radford2021learning}, and LPIPS~\cite{zhang2018unreasonable} as quantitative metrics.
During evaluation, the ViPer proxy model~\cite{salehi2024viper} is conditioned on 20 context images, comprising both preference and non-preference samples from the test set, and is then used to score the image generated by models.
To calculate win rate, for each prompt, we compare our output against the baseline FLUX.1-dev output, the image with the higher ViPer score is deemed the winner.
CLIP score employs CLIP~\cite{radford2021learning} to assess text-to-image alignment between the generated image and the input prompt, where a higher CLIP T2I score indicates better consistency. 
LPIPS~\cite{zhang2018unreasonable} measure the perceptual similarity between generated images and the user's preferred images under the same prompt, with lower LPIPS scores indicating closer adherence to user preferences.

\begin{figure}[!t]
    \centering
    \includegraphics[width=\linewidth]{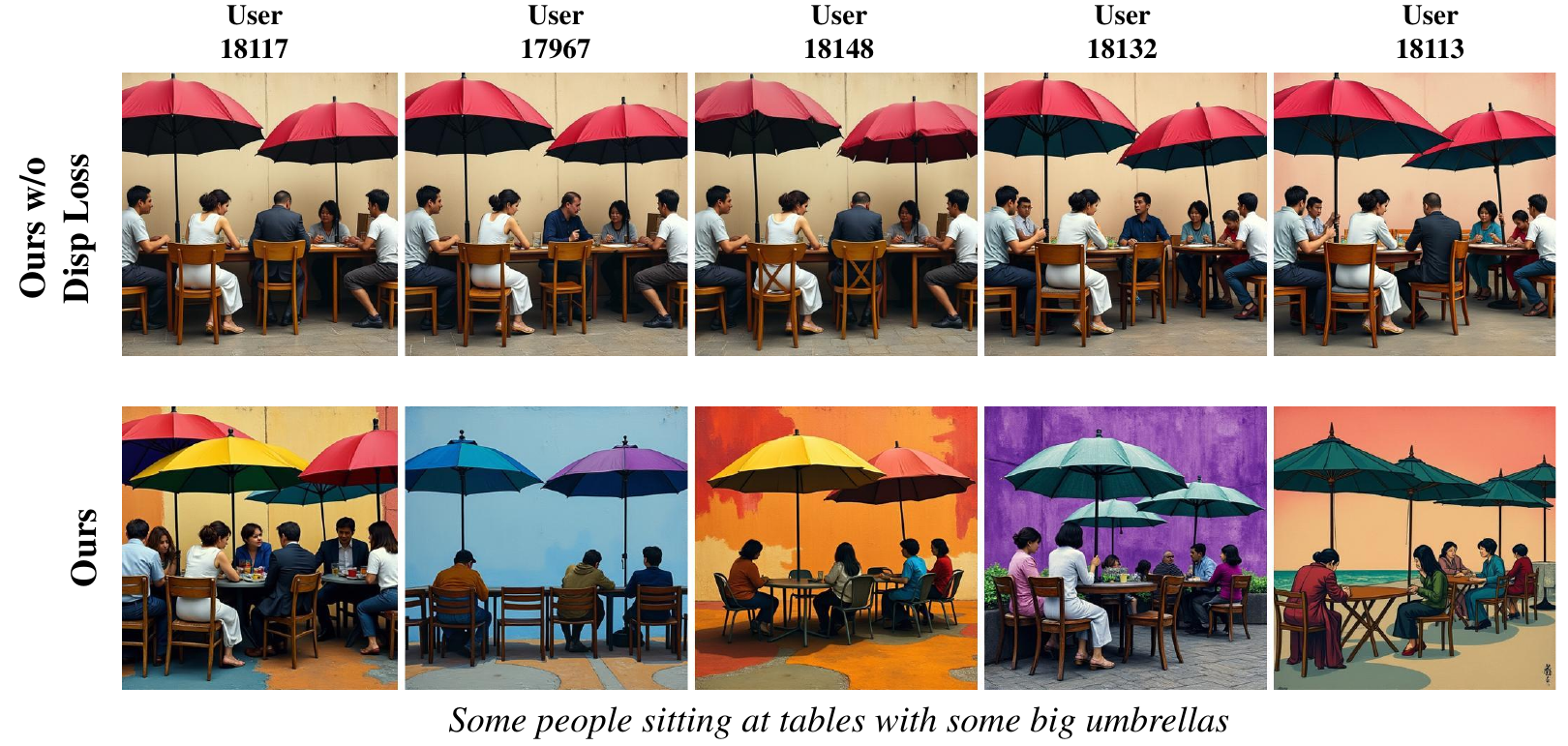}
    \caption{\textbf{Qualitative  dispersion loss ablation comparison of our method.} After ablating the dispersion loss, the generated preference images across different users exhibit substantially reduced variation. }
    \label{fig:result_disp}
\end{figure}

\subsection{Preference Alignment Comparison}

To demonstrate the effectiveness of our proposed method, we compare it with existing methods, including InstantStyle~\cite{wang2024instantstyle}, Bagel~\cite{deng2025emerging}, Qwen-Image-Edit~\cite{wu2025qwenimagetechnicalreport}, DrUM~\cite{kim2025draw}, and ViPer~\cite{salehi2024viper}.
Our method trains on eight historical preference images per user and adopts the strategy of learning linear combination coefficients.
Because DrUM and ViPer rely on textual inputs, we use JoyCaption~\cite{joycaption2024} to generate detailed captions from each user's historical preference images and provide these captions as comments or history text to the respective models.
Both models also receive the same set of eight preference images. 
For Bagel, we supply a single preference image and instruct the model to transfer its style, color, and other visual attributes to the newly generated image.
For Qwen-Image-Edit, which supports up to three input images, we provide a mix of preference and non-preference images and prompt the model to identify their differences and transfer the desired attributes, such as style and color, into the new image. 
InstantStyle is given a single user preference image as its reference.
As shown in~\cref{fig:result_compare} and ~\cref{tab:compare}, our method achieves superior performance under ViPer evaluation compared to other methods, while also attaining higher text-image alignment and perceptually closer resemblance to the user's preference images.
Our method produces images that are closer to user preferences in terms of style, color tone, geometric elements, and composition.
More experimental comparisons and performance analyses will be provided in the supplementary material.

\begin{figure}[!t]
    \setlength{\intextsep}{5pt}
    \setlength{\textfloatsep}{5pt}
    \setlength{\abovecaptionskip}{3pt}
    \setlength{\belowcaptionskip}{3pt}
    \centering
    \includegraphics[width=\linewidth]{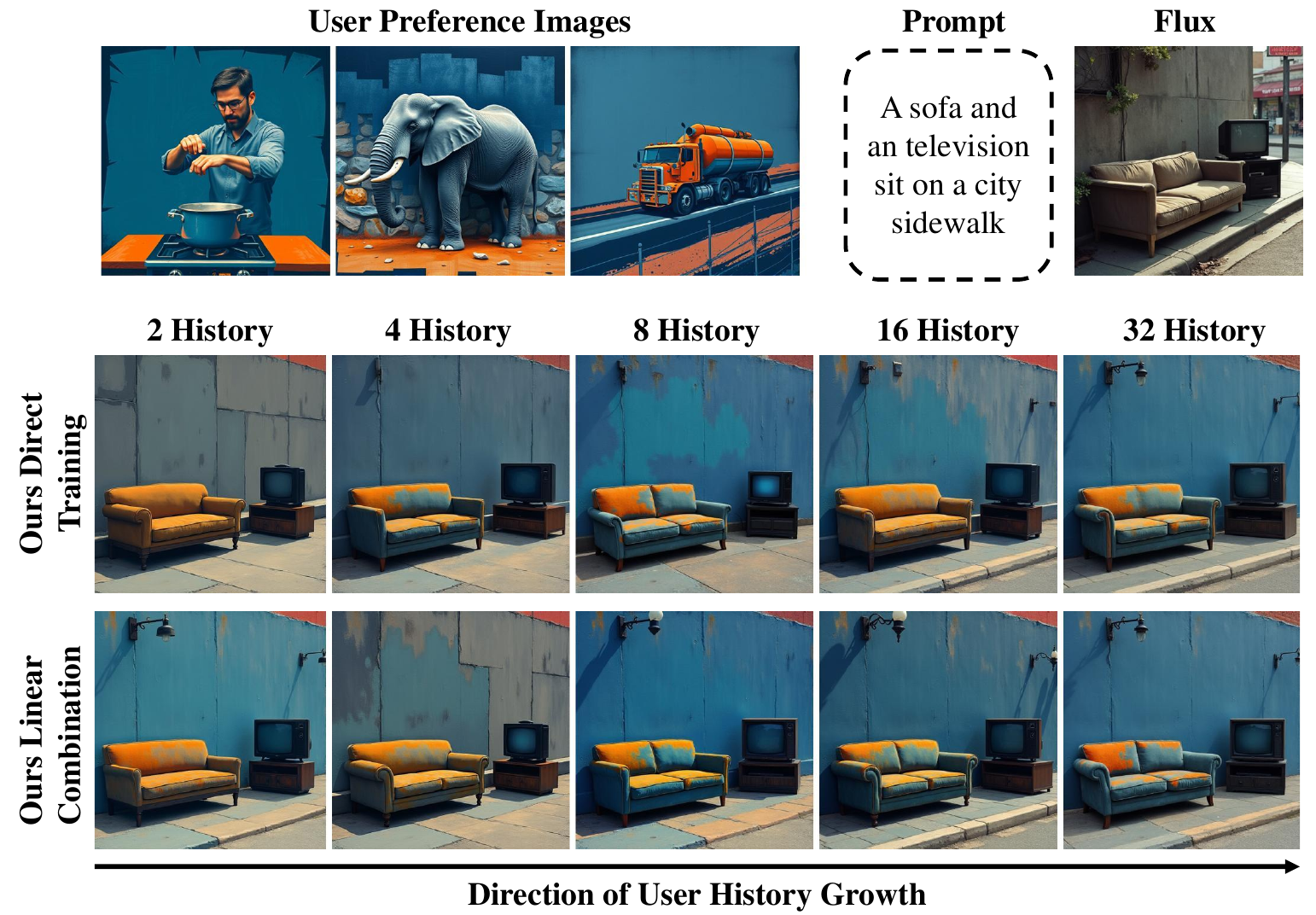}
    
    \caption{\textbf{Qualitative comparison of our method across different user history lengths and different training strategy} When the amount of user history is limited, training linear combination coefficients yields more stable performance.}
    \label{fig:result_history}
\end{figure}

\noindent \textbf{User Study.}
We also conduct the user study to assess our method from a human perceptual view.
For each test case, we first show human experts six images that the user have previously liked to establish the user's preference profile. 
We then perform an A/B test by presenting two images generated from the same text prompt: one produced by our method and the other by a competing baseline. 
Evaluators are asked to choose the image that better aligned with the user’s preferences while remaining more faithful to the given prompt. 
We collected responses from 40 human experts, and the aggregated results are reported in \cref{fig:human_evaluation}. 
As shown, our method is preferred by human experts, demonstrating superior alignment with both user preferences and text prompts compared to competing approaches.

\subsection{Ablation Study}

We have conducted a series of ablations to assess the effect of each proposed component.

\noindent \textbf{Effect of Linear Combination.}
To assess the impact of the linear combination strategy, we compare directly training user preference embeddings against optimizing linear combination coefficients of training-set embeddings under varying user history lengths (\ie, 2, 4, 8, 16, and 32). 
The experimental setup mirror those used in the main experiments.
As shown in the ~\cref{fig:history_viper_score} and ~\cref{fig:history_lpips} , when the history length is below 8, optimizing linear combination coefficients clearly outperforms training user embeddings from scratch. 
As more history becomes available, directly training user embeddings yields stronger quantitative results. 
However, the perceptual differences remain subtle, as illustrated in~\cref{fig:result_history}.
These findings suggest that representing new users as linear combinations of training-user embeddings is generally more effective for limited history scenarios.

% We investigate the performance of directly training user preference embeddings versus training linear combination coefficients under varying lengths of user history.
% %
% User history lengths are divided into 2, 4, 8, 16, and 32.
% %
% The experimental setup and test data for training user embeddings are identical to those used in the main experiments.
% %
% As shown in the ~\cref{fig:history_viper_score} and ~\cref{fig:history_lpips} , when the user history length is less than 8, training linear combination coefficients demonstrates a clear advantage over directly training the user preference embedding.
% %
% As the amount of user history increases, directly training the user preference embedding yields better quantitative results. However, the visual differences in the generated images remain subtle, as illustrated in the ~\cref{fig:result_history}.
% %
% This indicates that representing new users via linear combinations of training-set user preference embeddings better aligns with user preferences in the majority of cases.

% 
\noindent \textbf{Effect of Preference Adapters.}
We then ablate key components of the preference adapters using the same setup as the main experiments, with new user embeddings learned via linear combination optimization.
And the comparison results are illustrated in ~\cref{tab:abla} and ~\cref{fig:result_abla}.
Ours w/o $\Delta_\text{shared}$ denotes removing the block-shared modulation direction during preference adapter training.
Ours w/o $\Delta_\text{distinct}$ denotes removing the block-distinct modulation direction.
The results indicate that both the block-shared and block-distinct adapters are essential for effective preference modulation, while removing either one leads to degraded performance.

\noindent \textbf{Effect of Dispersion Loss.}
We also perform an ablation study on the introduced dispersion loss.
When this loss is removed, the generated images s across different users exhibit reduced diversity, confirming that the dispersion loss indeed enhances user distinguishability, as shown in ~\cref{fig:result_disp}.

\noindent \textbf{Effect of Prompt Preference Modulation.}
We final assess prompt preference modulation by replacing the user input text with an empty prompt, denoted as ``Ours w/o PPM''. 
As shown in~\cref{tab:abla}, removing prompt preference modulation results in a measurable decline in generation performance, demonstrating its effectiveness.

%% file: sec/6_conclusion.tex
\section{Conclusion}

Our method trains learnable user embeddings on user preference data to encode image preferences, and applies prompt preference modulation to obtain a more accurate and fine-grained preference modulation of user preferences.
The introduction of the dispersion loss enables our method to significantly enhance both individual preference discriminability and preference alignment.
To mitigate instability in preference generation when user history is limited, we represent new user preferences as combinations of well-trained embeddings from the training set.
Experimental results demonstrate that our approach achieves superior performance under both ViPer proxy evaluation and human expert assessment, while also generating images that better align with the user’s input text.
\begin{flushleft}
\noindent \textbf{Acknowledgement.}
The work was supported by National Natural Science Foundation of China under Grant No. 62371164.
\end{flushleft}

%% file: sec/X_suppl.tex
\clearpage
\setcounter{page}{1}
\setcounter{section}{0}
\setcounter{figure}{0}   % 重置图片计数器
\setcounter{table}{0}    % 重置表格计数器
\renewcommand\thesection{\Alph{section}}
\renewcommand\thesubsection{\thesection.\arabic{subsection}}
\renewcommand\thefigure{\Alph{figure}}
\renewcommand\thetable{\Alph{table}}
\renewcommand\thetable{\Alph{table}}
\renewcommand\theequation{A.\arabic{equation}}
\maketitlesupplementary

The following materials are provided in this supplementary file:
\begin{itemize}
    \setlength{\itemsep}{2pt}
    \setlength{\parsep}{0pt}
    \setlength{\parskip}{0pt}
    \item Sec.~\ref{sec:pip_data}: Comparison of Premier with other methods in the PIP dataset evaluation.
    \item Sec.~\ref{sec:generalization}:  Generalization performance on real user preference images.
    \item Sec.~\ref{sec:lora_comp}: Comparison with LoRA in terms of effectiveness and efficiency.
    \item Sec.~\ref{sec:other_analysis}: Additional experimental analyses, including hyperparameter sensitivity analysis and training-set user scale analysis.
    \item Sec.~\ref{sec:more_qualitative}: More qualitative results.
    
\end{itemize}
\section{Experimental Comparison on PIP-dataset}
\label{sec:pip_data}
\noindent \textbf{Evaluation Setup.}
Our method and several baseline approaches were evaluated on 20 randomly selected users from the PIP dataset, adhering to the experimental protocol established in the main paper.
For each user, we generated preference images using Z-Image and obtained the user preference embedding via the linear combination approach.

\begin{figure*}[!t]
    \centering
    \includegraphics[width=\linewidth]{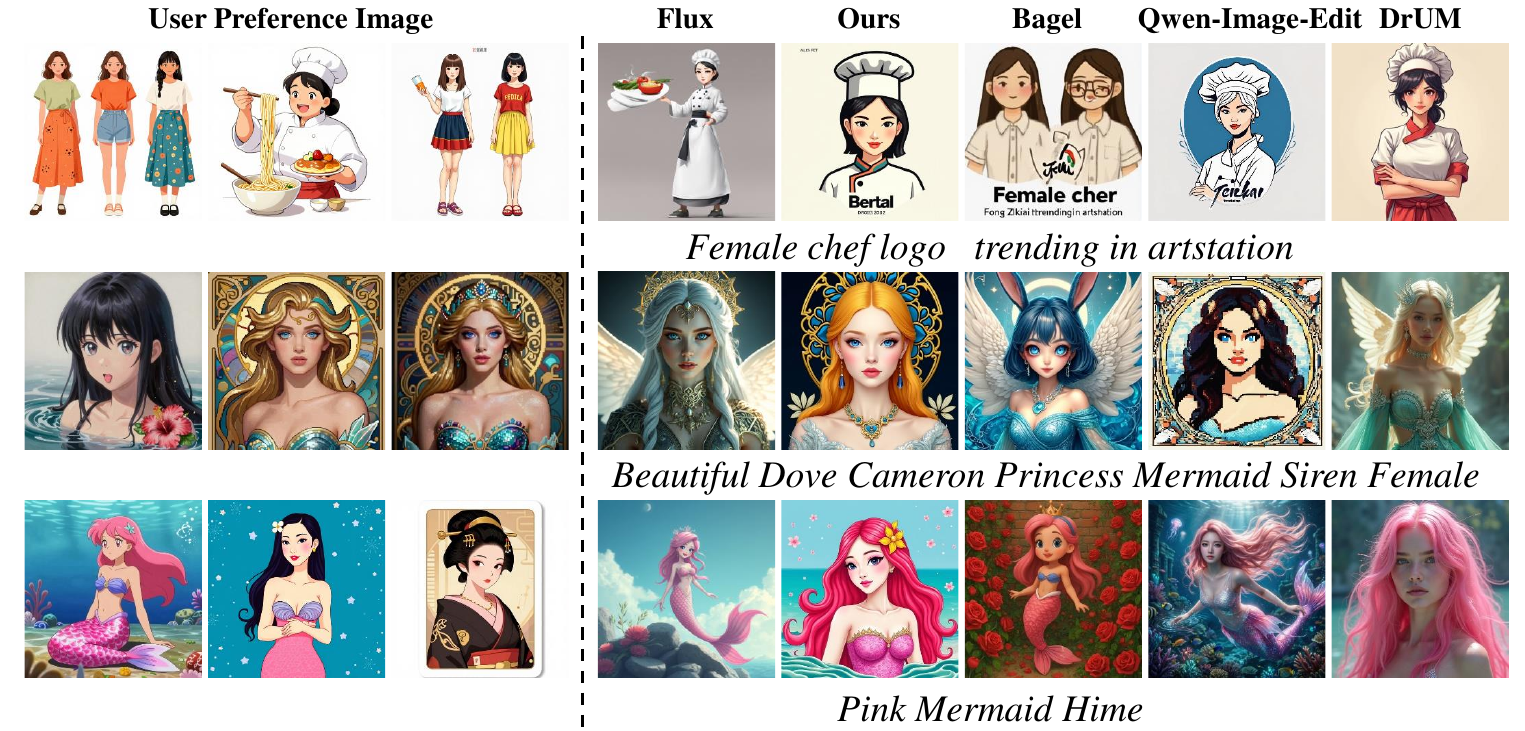}
    \caption{\textbf{Qualitative comparison on PIP-dataset.} }
    \label{fig:premier_rebuttal}
\end{figure*}

\noindent \textbf{Evaluation Results.}
In the PIP dataset, a single user's preferences often exhibit considerable diversity.
Our approach leverages user preference embeddings to enable token-level preference modulation, thereby adding user preferences more precisely.
From~\cref{tab:pip_compare}, our method achieves the best performance in terms of ViPer Score, ViPer Rate, and LPIPS.
From~\cref{fig:premier_rebuttal}, the images generated by our method are closer to the user's preferences.
\section{Generalization Performance}
\label{sec:generalization}
\noindent \textbf{Evaluation Setup.}
To evaluate the effectiveness of our method on real user data beyond the dataset, we select 4 real-world user preference cases and obtain their user preference embeddings for testing.
4 real-world user preferences are represented by the following: Honor of Kings, Arknights, Genshin Impact and Chiikawa.
For each preference, we select 8 preferred images as the user’s historical data and directly train the user embedding to obtain the new user’s preference embedding, enhancing generalization to out-of-distribution preferences.

\begin{figure*}[!t]
    \centering
    \includegraphics[width=\linewidth]{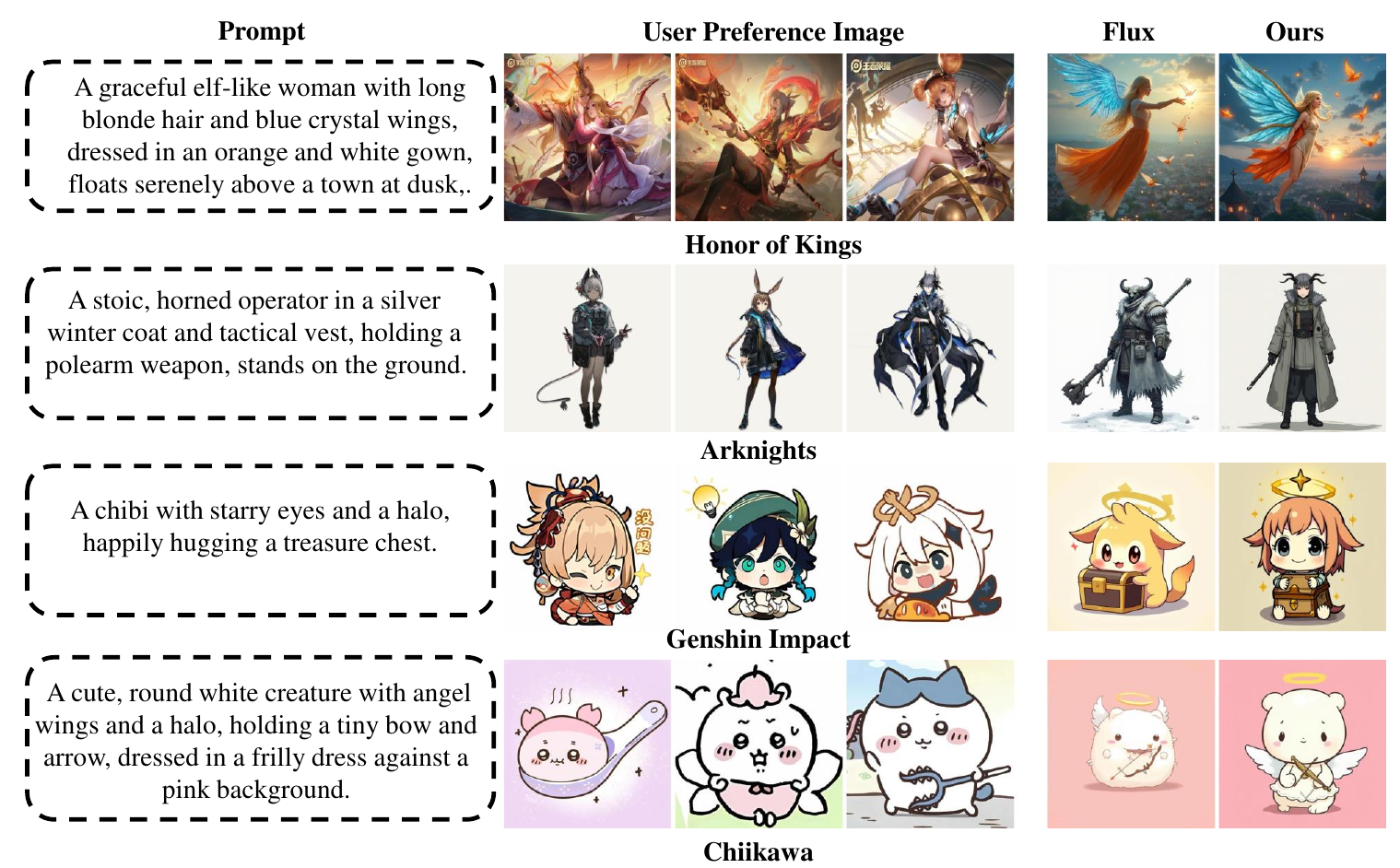}
    \caption{\textbf{Qualitative results on real user preference data.} }
    \label{fig:real}
\end{figure*}

\noindent \textbf{Evaluation Results.}
As shown in ~\cref{fig:real}, even on real-world preference data outside the training dataset, our method successfully captures key aspects of user preferences.
The characters generated by our method closely resemble the target characters in the user’s reference images. 
Specifically, for the \textit{Honor of Kings} preference, our method produces images that better align with the user’s favored 3D-anime aesthetic.
For the \textit{Arknights} preference, the outputs more faithfully reflect the stylistic conventions of 2D anime characters.
These results demonstrate the generalization capability of our approach to real-world user preferences.

\section{Comparison with LoRA}
\label{sec:lora_comp}
\begin{figure*}[!t]
    \centering
    \includegraphics[width=\linewidth]{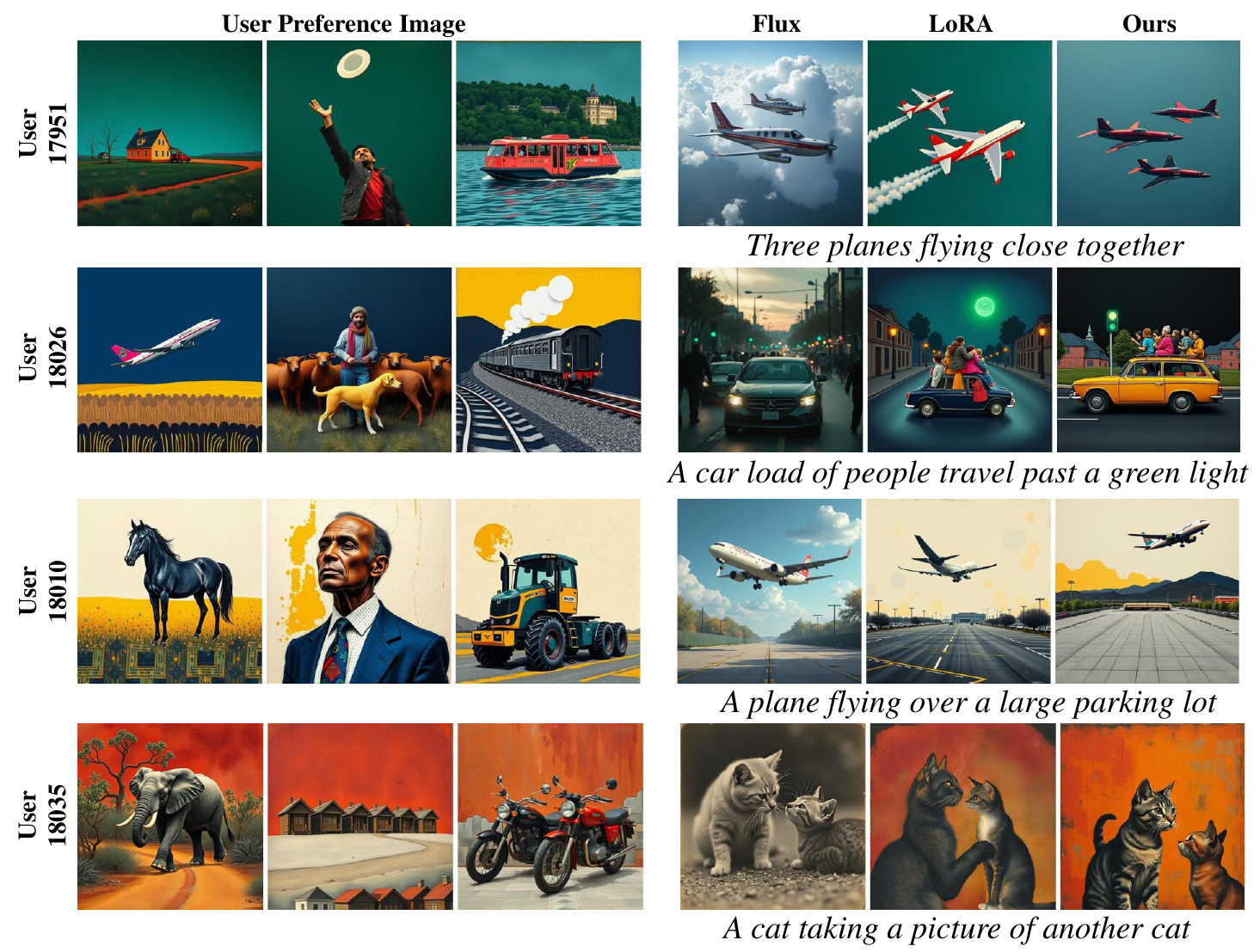}
    \caption{\textbf{Qualitative comparison with LoRA-generated results.} }
    \label{fig:lora}
\end{figure*}

\noindent \textbf{Evaluation Setup.}
In the comparison with LoRA, we train a LoRA adapter for each user on their 8 preference images, using the Prodigy~\cite{mishchenko2024prodigy} optimizer with a learning rate of 1.
The LoRA rank is set to 1, and all linear layers in the attention and feed-forward modules are targeted for adaptation, with training conducted for 5,000 steps.
We evaluate both methods using the same metrics as described in the main paper. The results are reported in ~\cref{tab:compare_lora} and \cref{fig:lora}.

\noindent \textbf{Evaluation Results.}
As shown in ~\cref{tab:compare_lora}, our method achieves performance comparable to LoRA in terms of preference alignment, while preserving the base model’s text-to-image alignment capability more effectively. 
~\cref{fig:lora} further shows that LoRA may unintentionally harm the base model’s ability.
In the first row of ~\cref{fig:lora}, the LoRA-generated image shows only two airplanes, failing to depict the “three airplanes” specified in the prompt.
And in the second row, it violates the instruction “passing through a green traffic light.” 
In contrast, our method remains more faithful to the input text.
The results indicate that our method has minimal impact on the generative capability of the base model.

\noindent \textbf{Efficiency Analysis.}
Moreover, our method is significantly more efficient: each user embedding occupies only 61 KB of storage, compared to 10.7 MB for LoRA.
Our preference adapters occupy 1.8 GB of storage in total.
When the number of users exceeds 170, our method requires less storage than LoRA.
Training takes 30 minutes per user, versus 1.2 hours for LoRA. 
During inference, our approach introduces only a 1-second overhead over the base model at any resolution, demonstrating high inference efficiency.

\input{table/lora_comparison}
\input{table/PIP}
\input{table/analysis}
\section{Other Analysis Experiments}
\label{sec:other_analysis}
\noindent \textbf{Training-set User Scale Analysis.}
To investigate the relationship between the size of the user training set and model performance, we trained the preference adapter on 100 randomly selected users who do not appear in the test set.
User embeddings are trained using linear combination approach, with all other settings identical to the evaluation setup described in the main paper.
As shown in the ~\cref{tab:analysis}, model performance significantly degrades as the number of users in the training set decreases. 
This also indicates that our method can accommodate a wider diversity of user preferences when trained with a larger user population.

\noindent \textbf{Hyperparameter Sensitivity Analysis.}
The sensitivity analysis on hyperparameters follows the same setup as in the main paper on 20 users of test set, with adjustments only $\lambda_\text{shared}$ and $\lambda_\text{distinct}$.
As shown in~\cref{tab:analysis}, the model is more sensitive to $\lambda_\text{shared}$  because $\Delta_\text{shared}$ affects preference modulation across all DiT blocks, making performance highly dependent on its associated loss term.
When $\lambda_\text{shared}$ is too large, the model prioritizes separating users over aligning with preferences, degrading performance.
When $\lambda_\text{shared}$ is too small, the dispersion loss becomes ineffective, also causing a significant performance drop.
The variation of $\lambda_\text{distinct}$ has a similar trend to that of $\lambda_\text{shared}$  but with a smaller impact, as 
$\Delta_\text{distinct}$ already exhibits strong inherent diversity.
\section{More Qualitative Results}
\label{sec:more_qualitative}
\begin{figure*}[!t]
    \centering
    \includegraphics[width=\linewidth]{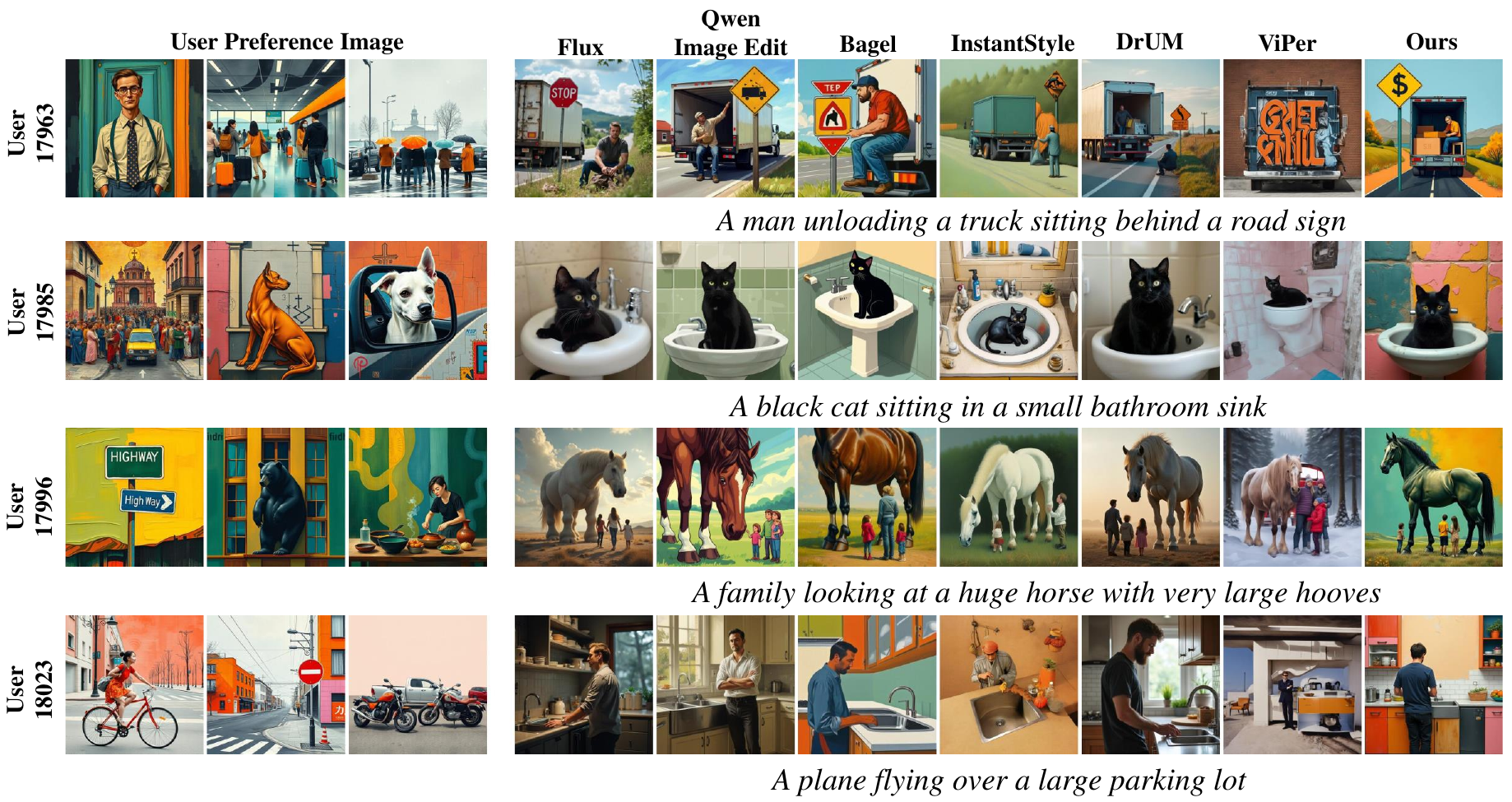}
    \caption{\textbf{Qualitative comparison with other methods.} }
    \label{fig:compare_0}
\end{figure*}

\begin{figure*}[!t]
    \centering
    \includegraphics[width=\linewidth]{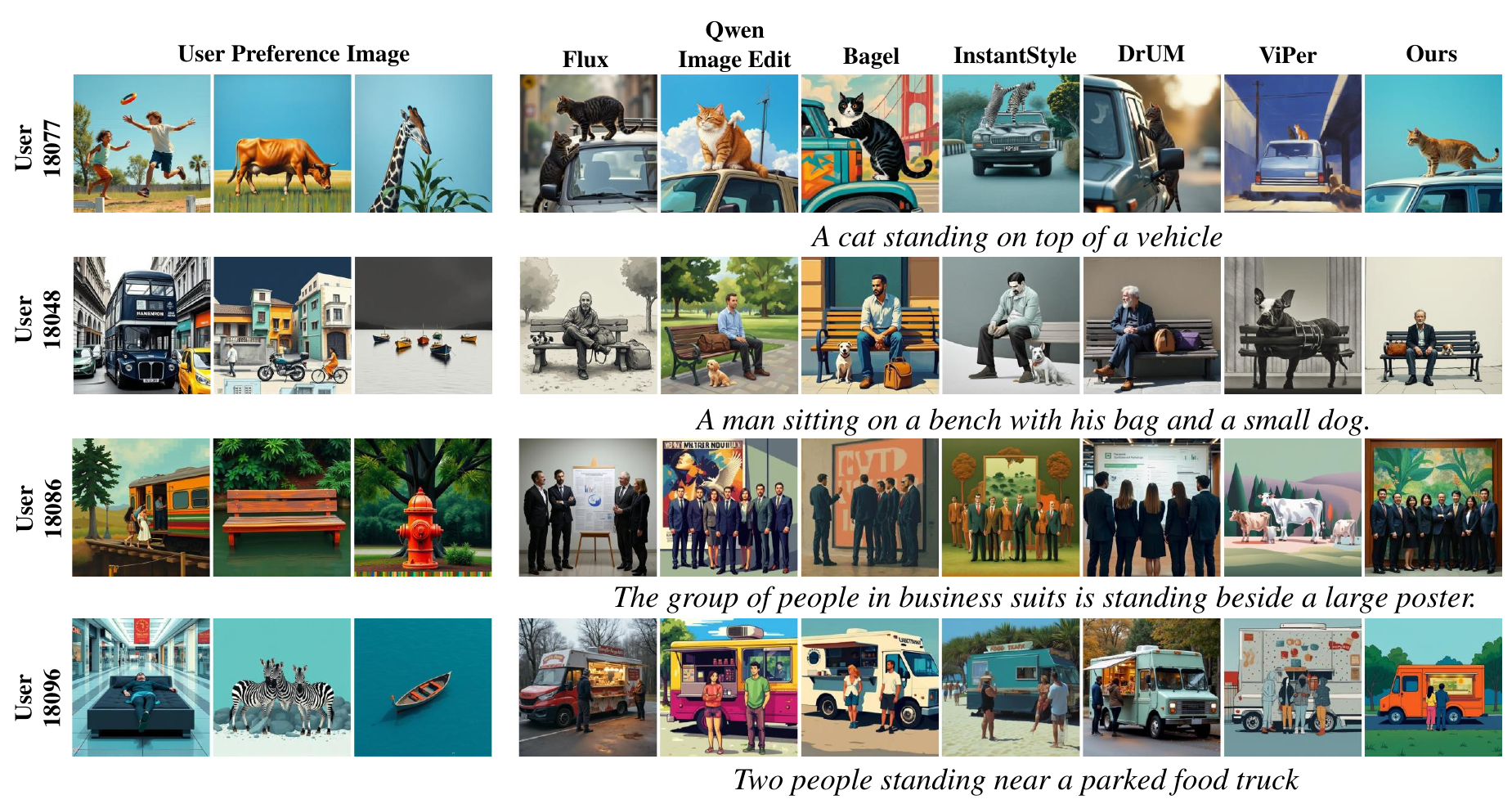}
    \caption{\textbf{Qualitative comparison with other methods.} }
    \label{fig:compare_1}
\end{figure*}

\begin{figure*}[!t]
    \centering
    \includegraphics[width=\linewidth]{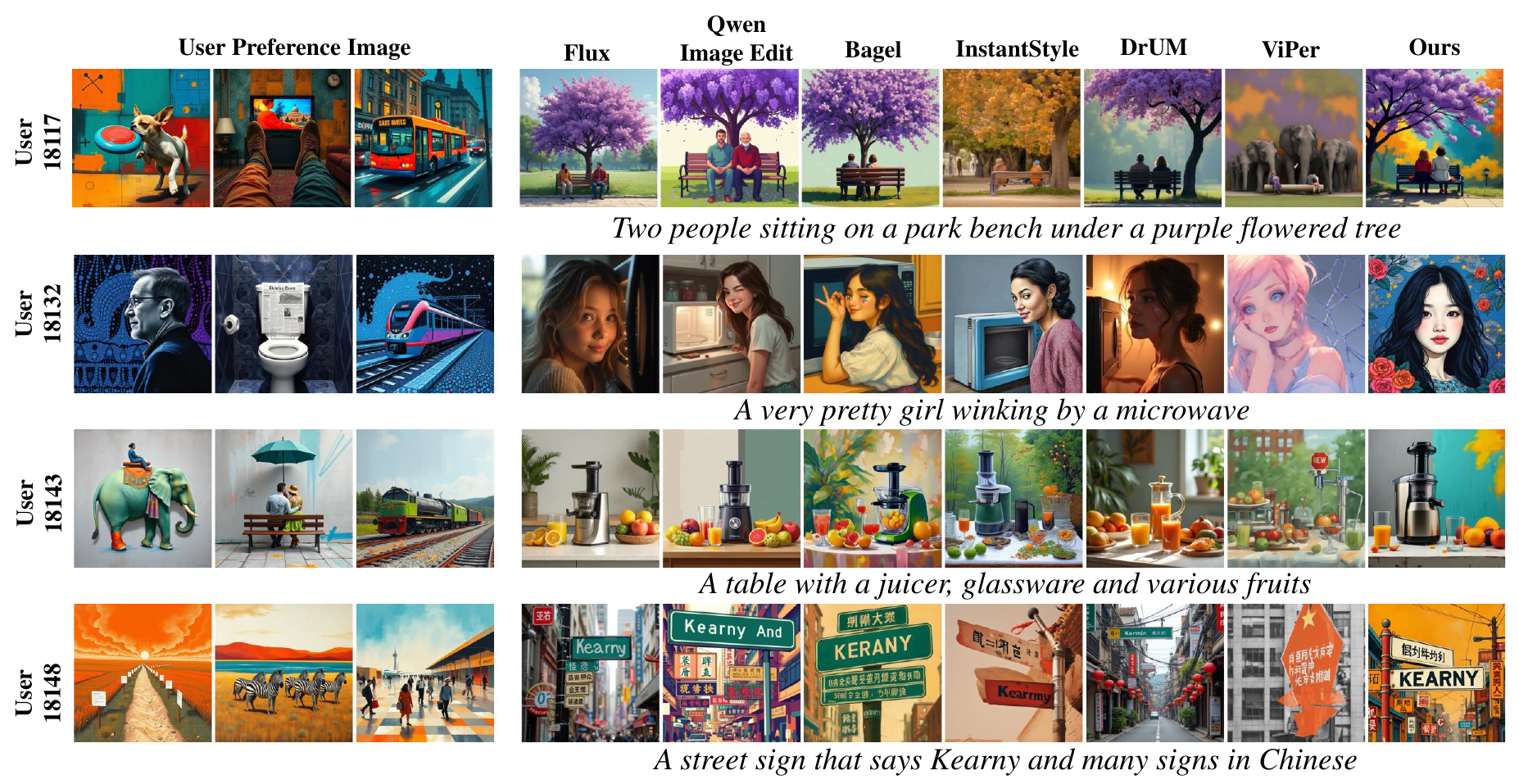}
    \caption{\textbf{Qualitative comparison with other methods.} }
    \label{fig:compare_2}
\end{figure*}

\begin{figure*}[!t]
    \centering
    \includegraphics[width=\linewidth]{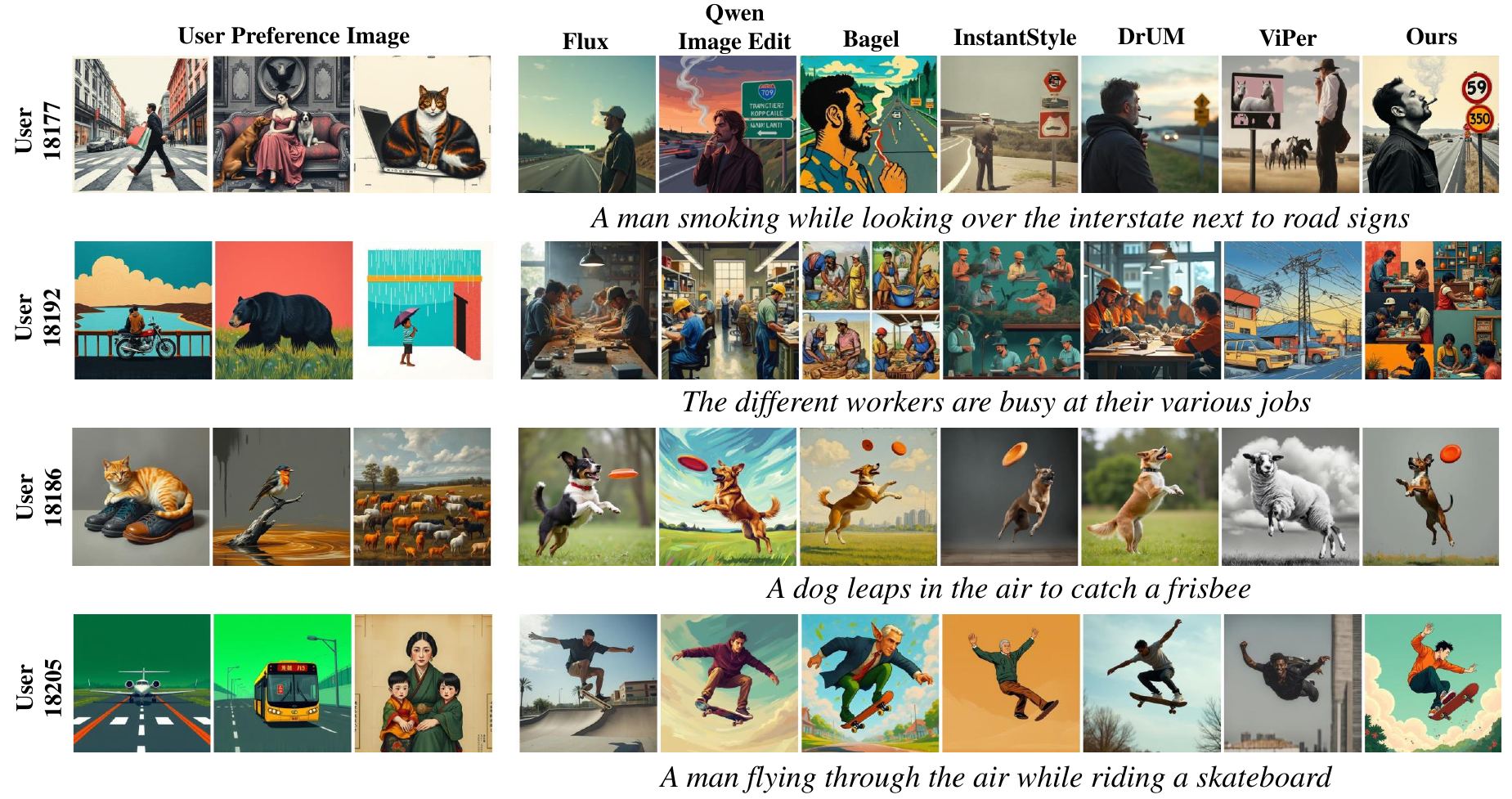}
    \caption{\textbf{Qualitative comparison with other methods.} }
    \label{fig:compare_3}
\end{figure*}

Additional qualitative comparisons with other methods are provided in ~\cref{fig:compare_0,fig:compare_1,fig:compare_2,fig:compare_3}.


%% file: table/lora_comparison.tex
\begin{table}
    \centering
    \resizebox{\linewidth}{!}{
        \begin{tabular}{l|cccccccc}\hline
        ~ & Data & Flux~\cite{labs2025flux1kontextflowmatching} & LoRA~\cite{liu2023flow} & Ours \\ \hline
        ViPer~\cite{salehi2024viper} Score$\uparrow$ & 0.8890 & 0.3953 & \textbf{0.7096} & 0.6889 \\ 
        ViPer Rate$\uparrow$ & - & - & \textbf{0.906} & 0.876 \\ 
        CLIP~\cite{radford2021learning} T2I$\uparrow$ & 0.3027 & 0.3089 & 0.3011 & \textbf{0.3183} \\ 
        LPIPS~\cite{zhang2018unreasonable}$\downarrow$ & - & - & 0.6037 & \textbf{0.5986} \\ \hline
    \end{tabular}
    }
    \caption{\textbf{Quantitative comparison  with LoRA training in terms of preference alignment in generated images.}  }
    \label{tab:compare_lora}
\end{table}

%% file: table/PIP.tex
\begin{table*}[t]
    \centering
    \resizebox{\linewidth}{!}{
        \begin{tabular}{l|ccccccc}
        \hline
        ~ & Data & Ours & Bagel & Qwen-Image-Edit & ViPer & InstantStyle & DrUM \\ \hline
        ViPer Score$\uparrow$ & 0.8919 & \textbf{0.7204} & \underline{0.7149} & 0.6696 & 0.5779 & 0.6232 & 0.6815 \\ 
        ViPer Rate$\uparrow$ & - & \textbf{0.6796} & \underline{0.6591} & 0.5399 & 0.4613 & 0.476 & 0.6134 \\ 
        CLIP T2I$\uparrow$ & 0.2844 & 0.2929 & 0.2653 & 0.2877 & 0.2823 & \textbf{0.2942} & \underline{0.293} \\ 
        LPIPS$\downarrow$ & - & \textbf{0.6002} & 0.6297 & \underline{0.6068} & 0.6306 & 0.643 & 0.6208 \\ \hline
        \end{tabular}
    }
    \caption{\textbf{Quantitative comparisons on the PIP-dataset.}}
    \label{tab:pip_compare}
\end{table*}

%% file: table/analysis.tex
\begin{table*}[t]
    \centering
    \resizebox{\linewidth}{!}{
        \begin{tabular}{c|cccccc}
        \hline
        ~ & Ours & Ours 100 users &  \makecell{Ours $\lambda_\text{distinct}=0.1,$ \\ $\lambda_\text{share}=0.01$}&  \makecell{Ours $\lambda_\text{distinct}=0.01,$ \\ $\lambda_\text{share}=0.1$} & \makecell{Ours $\lambda_\text{distinct}=0.1,$ \\ $\lambda_\text{share}=1$} & \makecell{Ours $\lambda_\text{distinct}=1,$ \\ $\lambda_\text{share}=0.1$} \\ \hline
        ViPer Score$\uparrow$& \textbf{0.7005} & 0.5013 & 0.5635 & 0.6632 & 0.6703 & \underline{0.6935} \\ 
        ViPer Rate$\uparrow$& \underline{0.8788} & 0.7115 & 0.7884 & 0.875 & 0.85 & \textbf{0.8807} \\ 
        CLIP T2I$\uparrow$& 0.3147 & \underline{0.3152} & \textbf{0.3153} & 0.3151 & 0.3139 & 0.3147 \\ 
        LPIPS$\downarrow$& \underline{0.5946} & 0.6261 & 0.6131 & 0.6014 & 0.6087 & \textbf{0.5923} \\ \hline
        \end{tabular}
    }
    % \textbf{Quantitative results of scaling analysis and sensitivity analysis for the proposed method in this paper.}
     \caption{\textbf{Results for scaling and sensitivity analysis.}}
    \label{tab:analysis}
\end{table*}